\documentclass{svproc}
\usepackage{graphicx}%
\usepackage{multirow}%
\usepackage{amsmath,amssymb,amsfonts}%
\usepackage{mathrsfs}%
\usepackage[title]{appendix}%
\usepackage{xcolor}%
\usepackage{textcomp}%
\usepackage{manyfoot}%
\usepackage{booktabs}%
\usepackage{algorithm}%
\usepackage{algorithmicx}%
\usepackage{algpseudocode}%
\usepackage{listings}%
\usepackage{url}

\begin{document}
\mainmatter              
\title{The Stigmergic Index: A Biomimetic Edge Centrality for Spatial Network Analysis}
\titlerunning{The Stigmergic Index: A Biomimetic Edge Centrality}  
%
\author{Sofiane Madmar\inst{1,2} \and Didier Josselin\inst{1} \and Vincent Labatut\inst{4}
\and Muhammad Zaly Shah\inst{3} \and Olivier Blight\inst{2}  }
\authorrunning{S. Madmar et al.} 
%


\institute{ESPACE, CNRS, Avignon University, Avignon, France\\
\email{sofiane.madmar@cnrs.fr}
\and
IMBE, Avignon Univ, Aix Marseille Univ, CNRS, IRD, Marseille, France\\
\and 
Centre for Innovative Planning and Development (CIPD), Faculty of Built Environment and Surveying, Universiti Teknologi Malaysia, Skudai 81310, Malaysia
\and
LIA UPR 4128,  Avignon University,  Avignon, France}

\maketitle              

\begin{abstract}
Ant Colony Optimization (ACO) is a family of metaheuristics inspired by the pheromone-based navigation of real ants, a decentralized mechanism often associated with shortest-path discovery. Over the past two decades, research has primarily focused on optimization tasks such as vehicle routing within a network. In this paper, we shift this perspective by using swarm intelligence to gather diagnostic information about the underlying network structure itself. Through decentralized exploration, we harness stigmergic dynamics to introduce a novel topological and behavioural metric: the Stigmergic Index. While sharing baseline structural properties with traditional betweenness centrality on simple grids, our index accounts for path polarization effects that can diverge from shortest-path-based assumptions.
\end{abstract}

\setcounter{footnote}{0}

\section{Introduction}\label{sec:intro}

Betweenness centrality is a fundamental concept in network science for identifying critical bottlenecks and estimating flow distributions. However, from the perspective of transportation geography, standard formulations rely on extreme and often unrealistic assumptions regarding navigation behaviour. On the one hand, geodesic betweenness~\cite{freeman_centrality_1991,brandes_faster_2001} assumes network-wide omniscience and strict rationality, routing all traffic exclusively along shortest paths while assigning no structural weight to viable alternative routes. Initially defined through nodes, it can also be extended to edges~\cite{Girvan_edge}. On the other hand, random-walk and current-flow betweenness measures~\cite{newman_measure_2005,brandes_centrality_2005} assume purely diffusive, undirected dynamics that lack goal-oriented efficiency. Real-world flows such as vehicular traffic operate between these two extremes, exhibiting \textit{bounded rationality}~\cite{mahmassani_bounded}---meaning that drivers choose a ``good enough" route instead of the perfect one due to limited time, information, and mental capacity. 

Swarm intelligence, and specifically Ant Colony Optimization (ACO) ~\cite{dorigo_optimization_1992,dorigo_ant_1999,dorigo_ant_2005}, offers a natural framework to embody this trade-off. Grounded in the foundational biology of collective foraging~\cite{deneubourg_self-organizing_1990}, ant colonies discover near-optimal paths through stigmergy: an indirect, self-organizing communication mechanism mediated by pheromone deposition and evaporation. This process inherently combines decentralized exploration with progressive positive feedback, allowing the colony to identify short paths without remaining trapped in a rigid, deterministic optimum. 

While ACO has been extensively investigated as a metaheuristic for solving combinatorial optimization and dynamic routing problems~\cite{blum_ant_2005,blum_ant_2024}, its emergent spatial dynamics have not been repurposed to diagnose network topology itself, to the best of our knowledge. Rather than viewing the pheromone trail merely as a computational means to find an optimal path, we propose to interpret the resulting pheromone landscape as an intrinsic, behaviourally based topological measure of the network. 

This paper introduces and explores the \textit{Stigmergic Index}, a novel biomimetic centrality measure proposed to bridge decentralized agent-based routing with structural spatial graph analysis. By simulating ant-inspired traffic across sampled Origin-Destination (O-D) pairs, our preliminary objective is to examine how this approach might account for alternative paths and topological redundancy that strictly geodesic metrics often overlook. As an initial validation, we benchmark this index against classical betweenness across selected synthetic topologies. Ultimately, this exploratory study aims to probe how the Stigmergic Index captures structural bottlenecks while conceptually integrating features of \textit{bounded rationality}, thereby outlining its potential as a behaviourally grounded alternative to deterministic centrality metrics.

The rest of the article is organized as follows. Section~\ref{sec:theory} details the theoretical foundations of ACO and our proposed methodology. In Section~\ref{sec:results}, we present the results of numerical experiments conducted on three models and compare them with betweenness centrality and random walk approaches. Finally, in Section~\ref{sec:discussion}, we discuss our novel proposition, its future developments and future insights.

\section{Theoretical Foundations and Methodology}
\label{sec:theory}

\subsection{Foundations of Ant Colony Optimization (ACO)}
\label{subsec:foundations}
Ant Colony Optimization \cite{dorigo_ant_1999} is a bioinspired metaheuristic that exploits the principle of stigmergy to solve combinatorial optimization problems on graphs. Let $G = (V, E)$ be an undirected network, where $V$ represents the set of vertices and $E$ the set of edges. A set of $K$ autonomous agents (ants) explores the graph topology to construct optimal paths between origin-destination (O-D) pairs. At each step, the probabilistic transition rule allowing an agent located at node $i$ to choose a neighbouring node $j \in \mathcal{N}(i)$ is formulated by combining the collective memory of the swarm with local heuristic information:
\begin{equation}
    P_{ij} = \frac{[\tau_{ij}]^\alpha [\eta_{ij}]^\beta}{\sum_{k \in \mathcal{N}(i)} [\tau_{ik}]^\alpha [\eta_{ik}]^\beta}, 
\end{equation}
where $\tau_{ij}$ measures the intensity of the pheromone trail on edge $(i,j)$. In classical ACO, $\eta_{ij}$ represents a static \textit{a priori} local heuristic, most commonly defined as the inverse of the edge weight, cost, or physical length ($\eta_{ij} = 1/c_{ij}$), with $c_{ij}$ the topological distance between node $i$ and $j$. The hyperparameters $\alpha \ge 0$ and $\beta \ge 0$ control the relative influence of the stigmergic memory and the local heuristic visibility, respectively. At the end of each iteration, pheromone trails undergo a global evaporation at a rate $\rho \in ]0, 1]$, followed by the deposition of new pheromones by agents that successfully reached their destination:
\begin{equation}
\begin{aligned}
    \tau_{ij}(t+1) = \max\left((1 - \rho)\tau_{ij}(t) + \sum \Delta\tau_{ij}^k, \, \tau_0\right), \\
    \text{with } \Delta \tau_{ij}^k = 
    \begin{cases} 
        \frac{Q}{L_k} & \text{if edge } (i,j) \in \mathcal{P}_k \\
        0 & \text{otherwise}
    \end{cases},
\end{aligned}
\end{equation}
where $Q$ is a positive deposition constant, $L_k$ is the total length of the path $\mathcal{P}_k$ traversed by ant $k$, and $\tau_0$ is the minimum pheromone concentration serving as a lower bound.

\subsection{A modification of the heuristic}
\label{subsec:heuristic}
During the stochastic process, in our first experiments, the standard ACO agents relying solely on link costs can perform ``blind" random walks that artificially increase pheromone levels along the boundaries of the graph. To mitigate this boundary effect, we change the heuristic information $\eta_{ij}$ from a static link cost to a geometric attraction slope directed specifically towards the destination node $D$ of the current O-D pair:
\begin{equation}
    \eta_{ij} = \frac{1}{d(j, D) + \epsilon}~,
\end{equation}
where $d(j, D)$ calculates the geometric Euclidean distance between the candidate node $j$ and the destination $D$, and $\epsilon > 0$ prevents division by zero. 

To optimize pathfinding in spatial graphs and avoid exploring divergent branches, using Euclidean distance as a heuristic function is a standard approach. Introduced by Hart et al. (1968)~\cite{hart_formal_1968} within the framework of the A* algorithm, it guarantees path optimality due to its admissibility. 

Although similar directional heuristics have been proposed in the literature for hybrid routing optimization, our contribution lies in integrating this spatial guidance into our biomimetic centrality metric (Stigmergic Index). This adaptation provides a more realistic model of biological rationality~\cite{madmar_unravelling_2026}, which tends to maintain a geometric heading, compared to a strictly topological calculation. 

While it may also include partly the need for omniscience of network structure in our algorithm, it is an essential modelling of the ants' ability to navigate based on visual and geometric cues of their environment, that leads to path optimization~\cite{Henquell198147}. Moreover, this spatial guidance mirrors the decision-making behaviour of human drivers, who naturally favour segments that maintain a general heading towards their destination while exploiting heavily trafficked routes.

\subsection{Proposed Approach: Stigmergic Index}
\label{subsec:stigmergy}
The primary objective of the proposed approach is to define a behaviourally grounded edge centrality metric that models emergent flow distributions within complex spatial networks. Unlike purely structural metrics that assume global network omniscience, our index models how local interactions, spatial headings, and positive feedback loops dynamically shape edge criticality. 

To achieve this, we define the \textbf{Stigmergic Index}. To compute the Stigmergic Index directly on the edges of a spatial network, 300 Origin-Destination pairs are sampled uniformly at random at each epoch. This fixed sampling budget functions analogously to mini-batch sizing in stochastic optimization: it provides a statistically representative evaluation of the network's structural variance while bypassing the prohibitive $O(|V|^2)$ computational cost of exhaustive sampling. For each selected O-D pair, a swarm of $K = 150$ ants explores the network. The simulation is executed over $N$ independent runs to account for stochasticity. In order to cover the network and iterate the process, we enforce a minimum of 5 epochs. Within each run, the global convergence of the pheromone landscape is evaluated at the end of each epoch by computing the cosine similarity $S_c$ between two successive edge pheromone vectors $\boldsymbol{\tau}^{(t)}$ and $\boldsymbol{\tau}^{(t-1)}$:
\begin{equation}
    S_c(\boldsymbol{\tau}^{(t)}, \boldsymbol{\tau}^{(t-1)}) = \frac{\boldsymbol{\tau}^{(t)} \cdot \boldsymbol{\tau}^{(t-1)}}{\|\boldsymbol{\tau}^{(t)}\| \|\boldsymbol{\tau}^{(t-1)}\|}.
\end{equation}
Convergence is reached once the stability condition $1 - S_c < \varepsilon_{\text{conv}}$ is satisfied, with a precision threshold set to $\varepsilon_{\text{conv}} = 10^{-2}$ (after a minimum of $5$ epochs). 

Let $\tau_e^{(r)}$ denote the stabilized pheromone concentration on edge $e \in E$ obtained at the end of run $r \in \{1, \dots, N\}$. We first compute the expected pheromone load across all runs:
\begin{equation}
    \bar{\tau}_e = \frac{1}{N} \sum_{r=1}^{N} \tau_e^{(r)}.
\end{equation}
The final Stigmergic Index $I_{\text{stig}}(e)$ is obtained by normalizing this distribution by the total network load:
\begin{equation}
    I_{\text{stig}}(e) = \frac{\bar{\tau}_e}{\sum_{e' \in E} \bar{\tau}_{e'}}.
\end{equation}
Under this formulation, $\sum_{e \in E} I_{\text{stig}}(e) = 1$, representing an empirical probability distribution of edge utilization that directly enables rigorous inequality analysis via Lorenz curves and Gini coefficients.

\subsection{Experimental Setup}
\label{subsec:setup}
We tested our approach on three spatial models: a $10 \times 10$ Manhattan grid, a Twin Cities network of two $8 \times 8$ grids with a dual-path bridge, and a 60-node random Delaunay network~\cite{Delaunay_1934aa}. In these initial experiments, we used unweighted and non-hierarchical networks. 

To evaluate the proposed Stigmergic Index, the simulation parameters were calibrated to reflect a realistic balance between collective traffic memory and individual spatial routing behaviour. For each randomly selected O-D pair, the swarm size is set to $K = 150$ ants. This population size ensures sufficient stochastic exploration of the network topology while maintaining scalability.

To rigorously calibrate the model, we conducted a preliminary sensitivity analysis on the Manhattan grid over the parameter space $\alpha \in \{0.5, 1.0, 2.0, 4.0\}$, $\beta \in \{0.5, 1.0, 2.0, 4.0\}$, and $\rho \in \{0.1, 0.3, 0.5, 0.8\}$. The simulation protocol was executed over $N = 3$ independent runs for each combination, deposition factor $Q = 10.0$, initial pheromone $\tau_0 = 0.1$, and convergence threshold $\varepsilon_{\text{conv}} = 10^{-2}$. For each configuration, we measured the average Spearman rank correlation~\cite{Spearman_1904} between the stabilized edge pheromone distribution and exact betweenness centrality to evaluate how the hyperparameters trade-off captures the network's structural hierarchy. 

Based on these findings, we set $\alpha = 1.0$, $\beta = 2.0$, and $\rho = 0.30$. Higher values ($\alpha \ge 2.0$) cause cyclic trapping and correlation collapse, while an excessive heuristic weight ($\beta = 4.0$) turns agents into deterministic followers. Finally, $\rho = 0.30$ balances memory retention and exploration, avoiding the path lock-in seen at $\rho = 0.10$.

\begin{figure}[h!]
    \centering
    \includegraphics[width=0.8\linewidth]{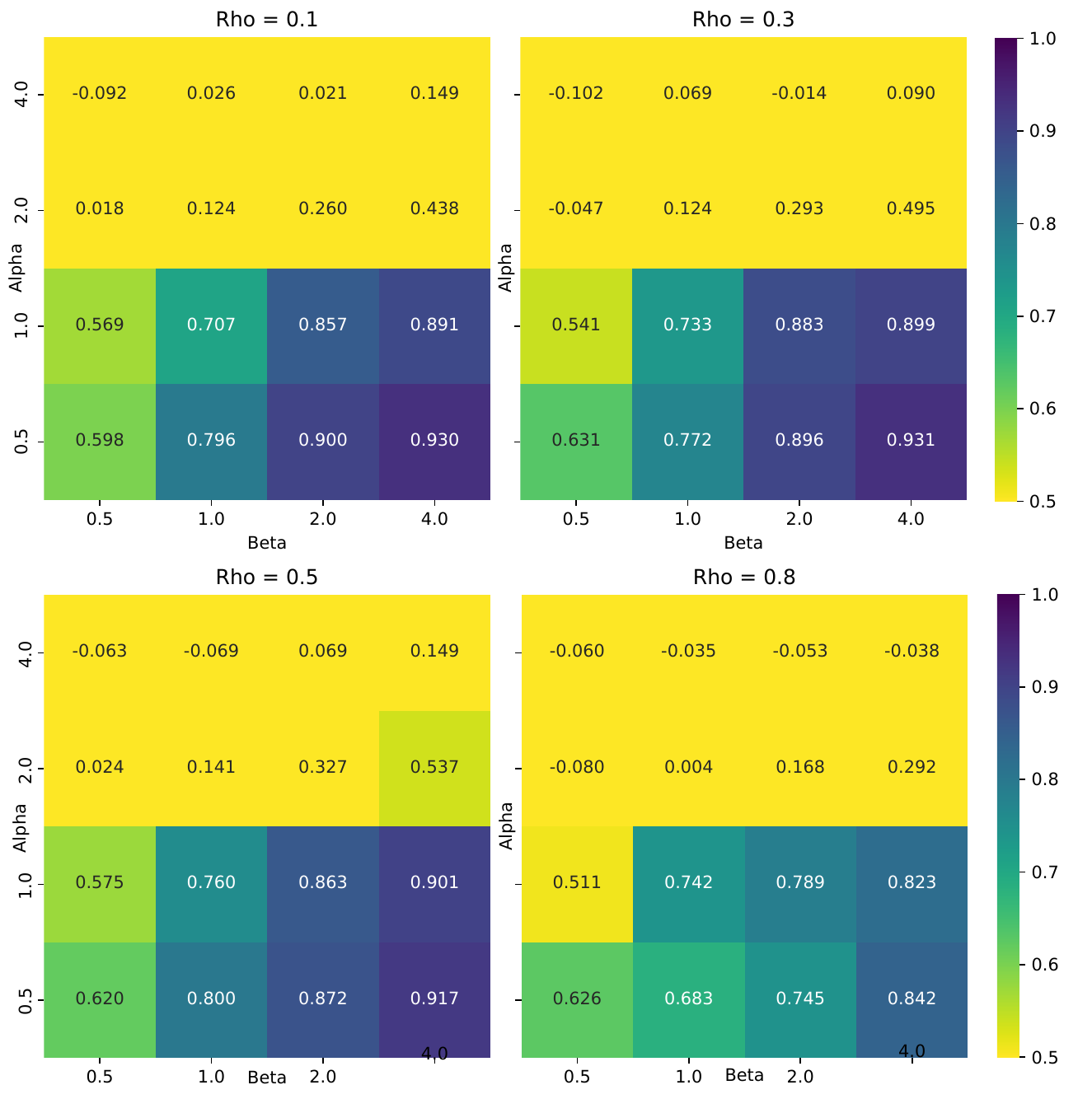}
    \caption{Sensitivity analysis of the Stigmergic Index on a $10 \times 10$ Manhattan grid. Heatmaps show the Spearman correlation between the stabilized pheromone distribution and betweenness centrality across combinations of $\alpha$, $\beta$, and $\rho$. $K=150$, $N=3$.}
    \label{fig:sensitivity}
\end{figure}

The probabilistic transition rule is governed by the pheromone weight $\alpha = 1.0$ and the spatial heuristic weight $\beta = 2.0$. This specific configuration ($\beta > \alpha$) is intentionally chosen to prioritize the geometric attraction slope over collective memory. It accurately models the cognitive behaviour of human drivers, who exhibit a strong tendency to maintain a direct geographical heading towards their target, using established traffic flows primarily as a secondary decision factor. 

Given the stochastic nature of our approach, we repeat the simulation $N=10$ runs in the experiments. The remaining simulation parameters are fixed to a deposition factor $Q = 10.0$, an initial pheromone level $\tau_0 = 0.1$, and a convergence threshold $\varepsilon_{\text{conv}} = 10^{-2}$.

We use the Spearman rank correlation to measure the proximity between our indicator and exact betweenness centrality from Brandes' algorithm. We also compare it to Newman's approach of random walk applied to networks. We use the respective functions from the \textit{NetworkX} Python library. We use the term `load' as a unified terminology to describe the value of each indicator.

\section{Results}
\label{sec:results}

First, the results show high Spearman correlation between betweenness and Stigmergic Index, synthesised in Table~\ref{tab:spearman}---which also references the related figures depicting each network. In the same table, we also report the Average Run, which presents the mean and standard deviation of the correlation with betweenness of edge pheromone rate calculated individually for each single run. Reporting the mean and standard deviation across these $N$ independent executions demonstrates that each unique run of the algorithm is highly stable and performant. The minimal variance observed from one execution to another confirms that the stochastic nature of the uniformly sampled O-D pairs does not compromise the algorithm's reliability. Instead, the dynamic pheromone accumulation consistently and robustly converges to the same structural criticality, regardless of the initial random sampling. 
\begin{table}[htpb]
\centering
\renewcommand{\arraystretch}{1.15}
\begin{tabular*}{\linewidth}{@{\extracolsep{\fill}}lccc}
\toprule
\textbf{Model} & \textbf{$I_{\text{stig}}$} & \textbf{Average Run} & \textbf{Figure} \\
\midrule
Manhattan            & 0.9703 & $0.8622 \pm 0.0108$ & Figure~\ref{fig:manhattan} \\
Modified Twin Cities & 0.8966 & $0.7627 \pm 0.0411$ & Figure~\ref{fig:twincities} \\
Delaunay Network     & 0.6180 & $0.5093 \pm 0.0461$ & Figure~\ref{fig:delaunay_log} \\
\bottomrule
\end{tabular*}
\caption{Spearman Correlation Coefficient Between Betweenness Centrality and Stigmergic Index Across the Benchmark Models, With Aggregated Mean Stigmergic Index or Mean of Average Run with Standard Deviation. $K=150$, $N=10$.}
\label{tab:spearman}
\end{table}

Furthermore, the Spearman rank correlation of the Stigmergic Index is consistently higher than the mean correlation of the individual runs. This demonstrates the clear benefit of aggregating pheromone concentrations across all $N$ executions, as this ensemble averaging effectively filters out residual stochastic noise and captures a more robust structural signal. 

In the grid network of the Manhattan model, our approach yields results very similar to betweenness centrality. Through iteration, agents explore the entire network, and central edges are clearly marked, indicating higher traffic volume. However, the modified Twin Cities model aims to test the stigmergic index's ability to highlight important edges that betweenness centrality fails to identify. This is precisely the rationale behind the Newman approach, which strongly marks both the direct edge connecting the two sides and the two-edge bridge. The stigmergic index, for its part, preserves the overall distribution of betweenness centrality while placing slightly more emphasis on the triangle top-bridge. It will be interesting to determine which ACO parameters can enhance this capacity for exploring and evaluating alternative central paths in a network.
The model with the lowest correlation coefficient is the Delaunay network which could inform us that the nature of Stigmergic Index is different from exact betweenness in more complex networks or in specific topologies.

\begin{figure}[h!]
    \centering
    \includegraphics[width=1\linewidth]{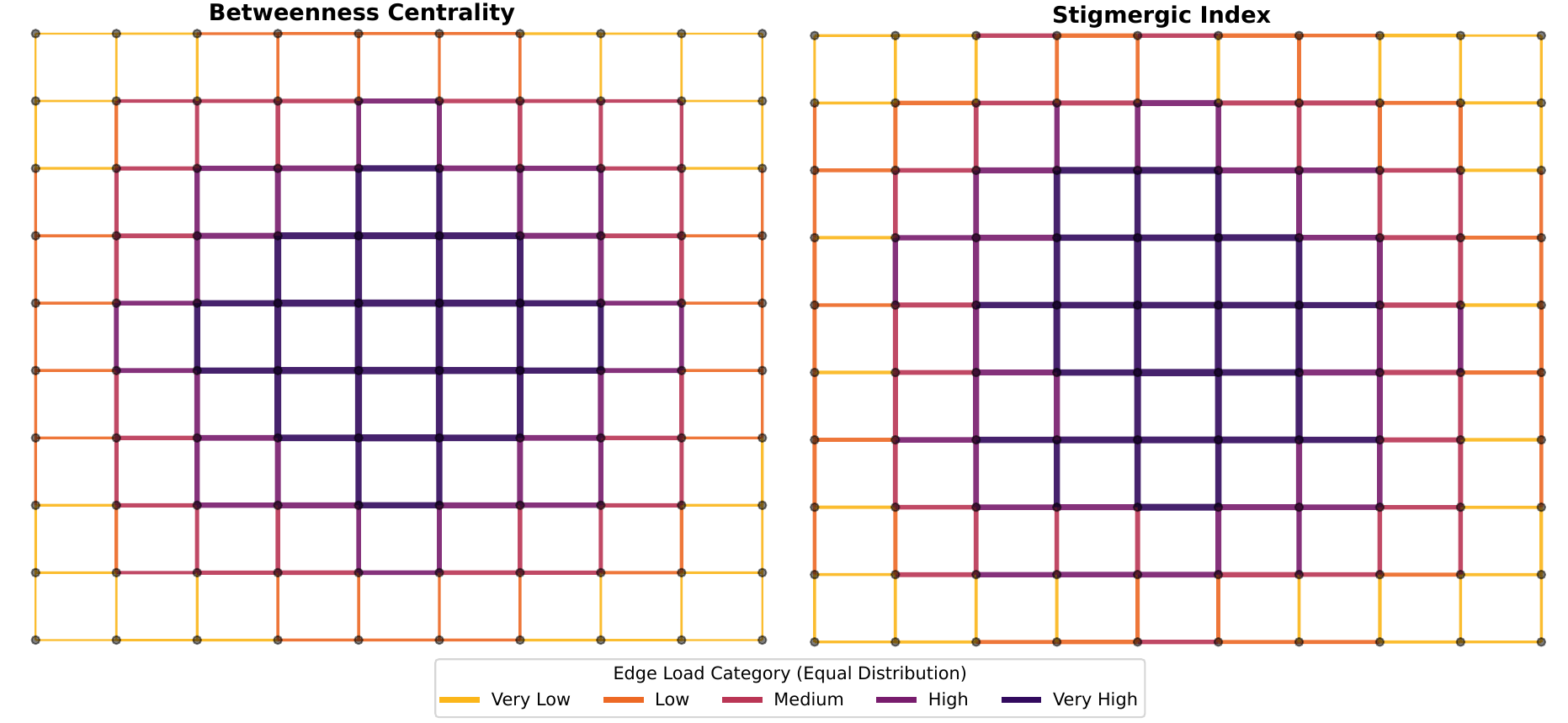}
    \caption{Load Repartition on Manhattan Grid using Betweenness Centrality and Stigmergic Index. $K=150$, $N=10$.}
    \label{fig:manhattan}
\end{figure}
\begin{figure}[h!]
    \centering
    \includegraphics[width=0.8\linewidth]{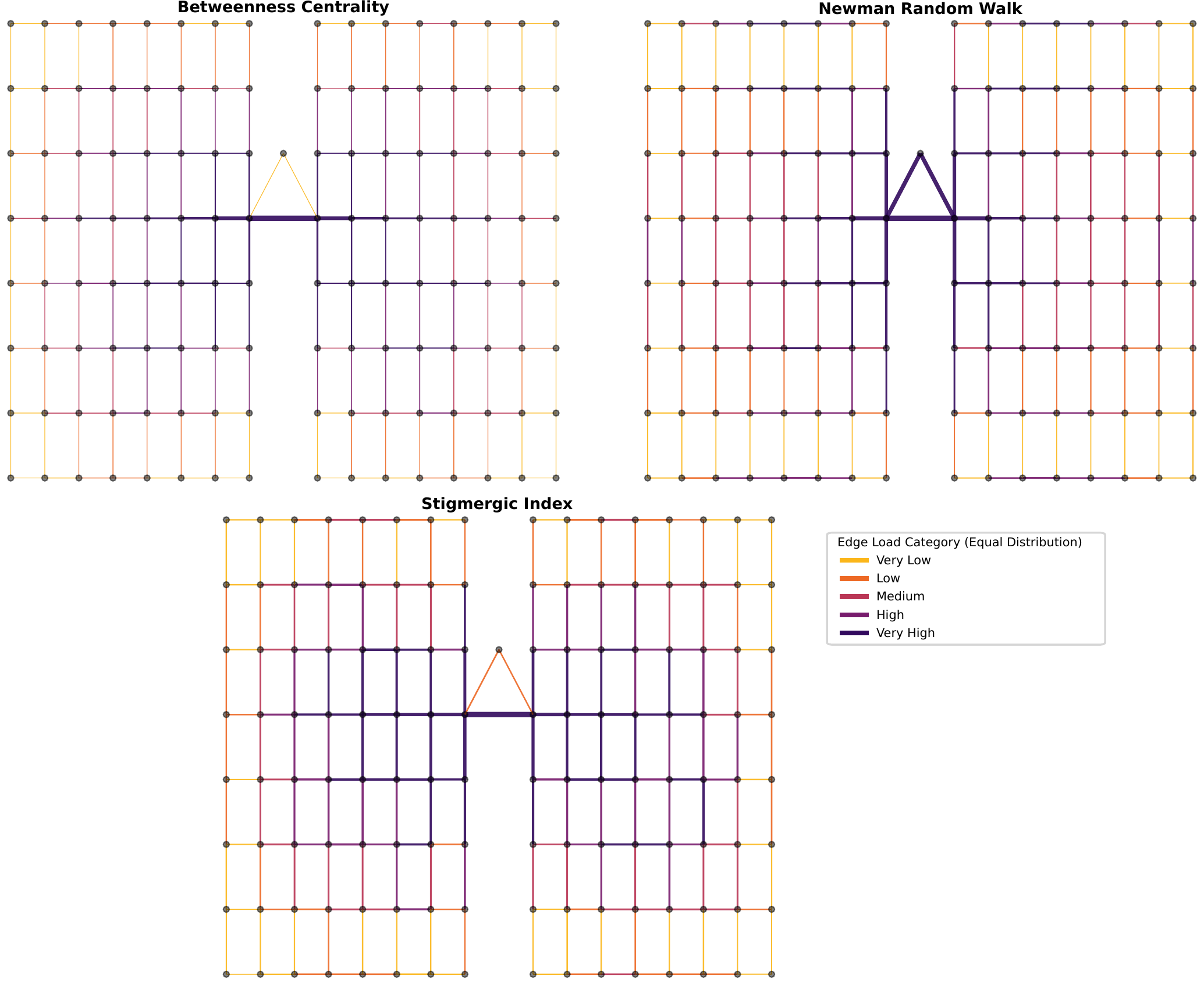}
    \caption{Load Repartition on Modified Twin Cities Model using Betweenness Centrality, Newman Random Walk  and Stigmergic Index. $K=150$, $N=10$.}
    \label{fig:twincities}
\end{figure}

\begin{figure}
    \centering
    \includegraphics[width=1\linewidth]{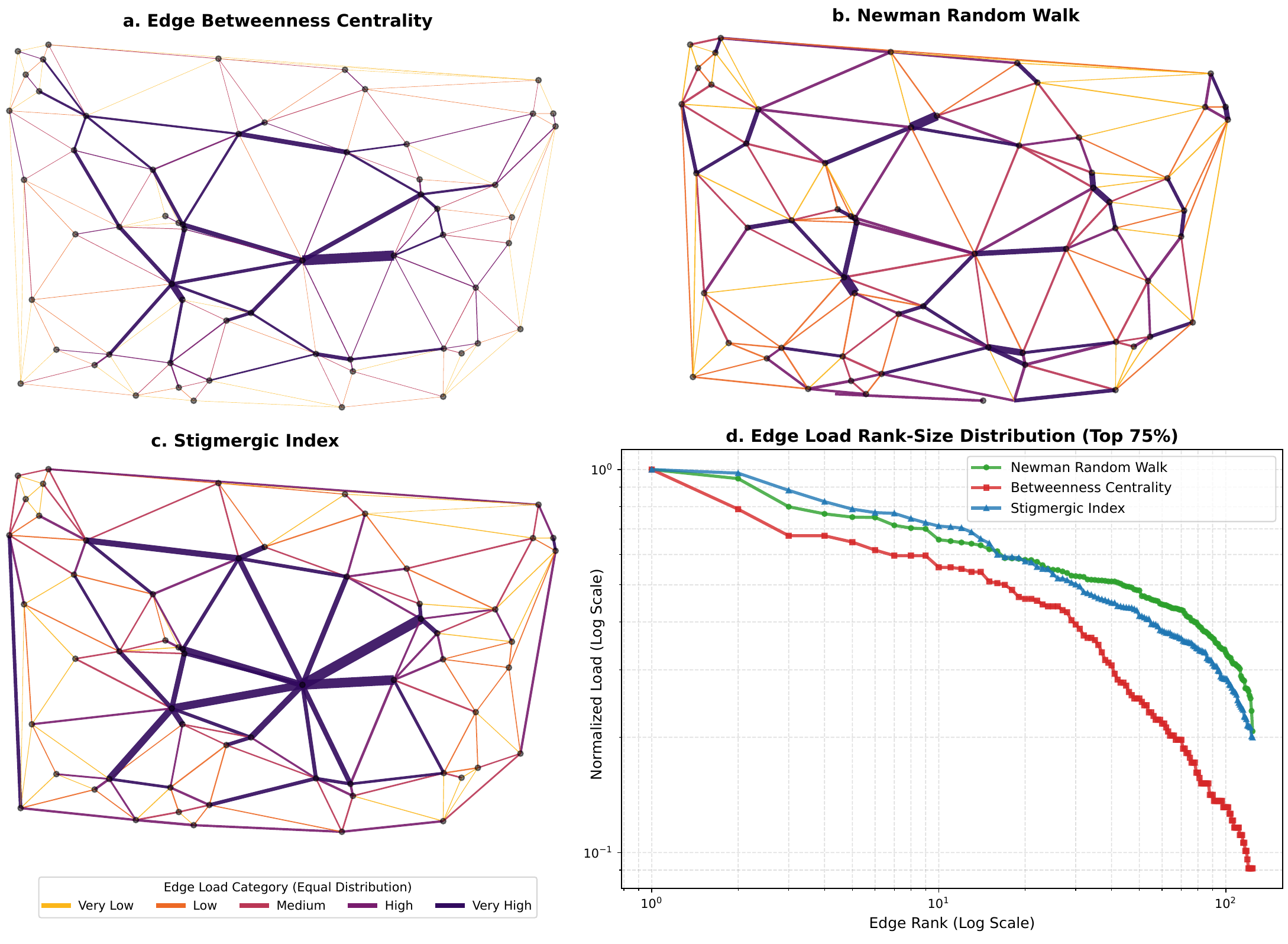}
    \caption{Load Repartition in a Delaunay Network using (a) Betweenness Centrality, (b) Newman Random Walk and (c) Stigmergic Index, with (d) an Edge Rank Size Distribution Comparison for Top 75\% Edges. Edge thickness on graphs represents the absolute load. $K=150$, $N=10$.}
    \label{fig:delaunay_log}
\end{figure}

\begin{figure}
    \centering
    \includegraphics[width=0.7\linewidth]{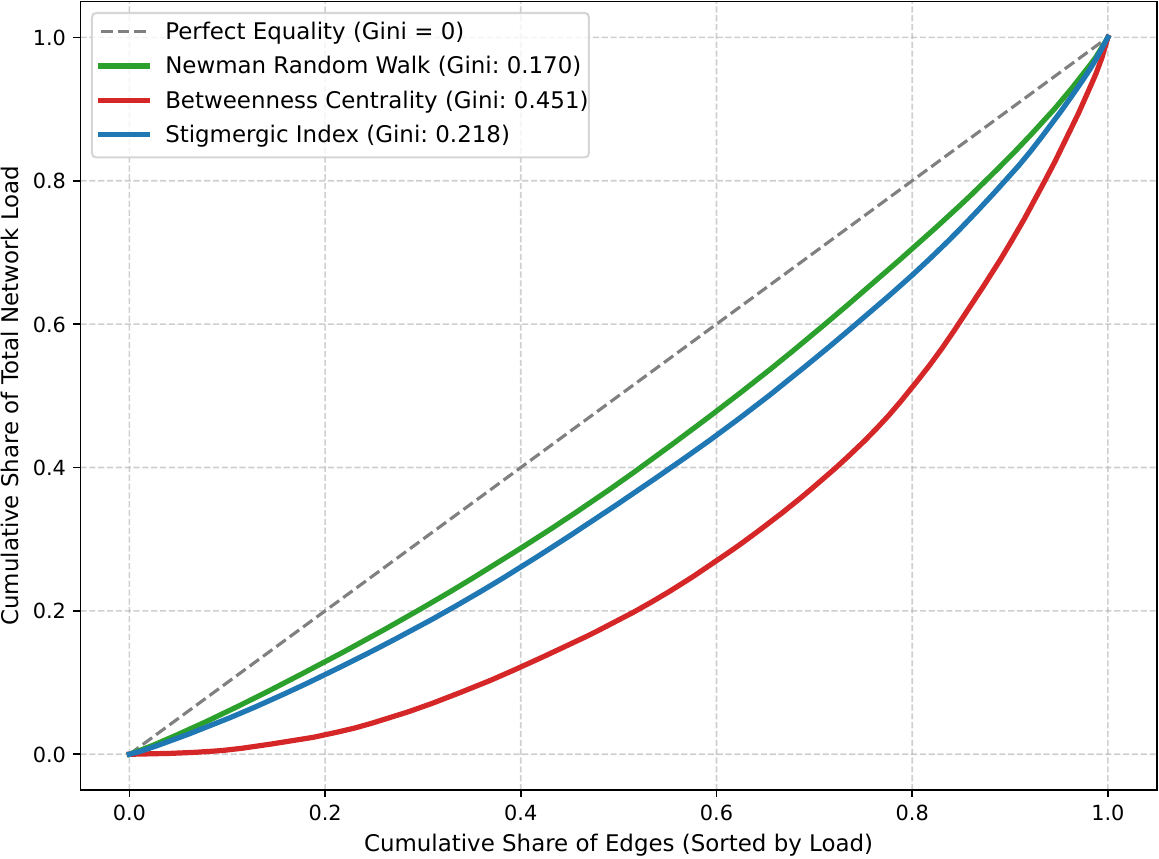}
    \caption{Lorenz curves and Gini coefficients comparing the three approaches. The dashed line represents perfect equality of network load. $K=150$, $N=10$.}
    \label{fig:delaunay_lorenz}
\end{figure}

In the Delaunay model, we observe a phenomenon of hyper-congestion in central edges (Figure~\ref{fig:delaunay_log}). The colony seems to abandon the exhaustive exploration of minor detours, forcing traffic to converge onto a restricted number of dominant axes. To quantify the structural differences between mathematical shortest-path routing, random diffusion, and our biologically-inspired mechanism, we analysed the inequality of load distribution across the Delaunay network using a Log-Log representation of top 75\% edges---to have a better view of the main edges---in Figure~\ref{fig:delaunay_log} and a Lorenz curve and its Gini coefficients~\footnote{The Lorenz curve graphically illustrates inequality by plotting the cumulative share of a resource (such as network load) against the cumulative share of entities (such as edges). The Gini coefficient, derived from this curve, quantifies this inequality into a single value ranging from 0 (perfect equality) to 1 (maximal inequality).} in Figure~\ref{fig:delaunay_lorenz}.

The results reveal a distinct hierarchical stratification among the three approaches. The Newman random walk centrality exhibits the lowest Gini coefficient at 0.170, reflecting a highly diffused, fluid-like load distribution where agents blindly disperse across all available topological detours. Conversely, the betweenness centrality yields the highest Gini coefficient at 0.451. This demonstrates the extreme polarization inherent to strictly geodesic routing: edges that do not lie on a strict shortest path are of lowest rank, creating a rigid and highly unequal load distribution. Crucially, the Stigmergic Index positions itself structurally and statistically between the two. Its Lorenz curve is significantly more equitable, yet more concentrated than Newman. This intermediate Gini coefficient at 0.218 illustrates how local probabilistic decisions and limited spatial awareness can account for non-deterministic route choices. Driven by heuristic parameters ($\alpha, \beta$), the majority of the swarm converges to the dominant macroscopic arteries; however, the stochastic transition rule ensures a continuous, low-intensity exploration of suboptimal secondary routes. This is then confirmed by the Log-Log analysis of load as a function of edge rank. In fact, it also positions our approach in between classic betweenness and Newman for most of edges (right side of the chart, lower load) while the highest edges concentrate a significantly higher share of the total load compared to them (left side of the chart, high load). The higher load concentration suggests a polarization mechanism due to the over-concentration of pheromones in the main parts of the networks as depicted visually (Figure~\ref{fig:delaunay_log}.d.).

\section{Discussion and Future Works}
\label{sec:discussion}

The previous results show that our approach may soften the unrealistic mathematical rigidity of exact shortest-path algorithms, producing a traffic load distribution that mirrors the imperfect, exploratory, yet goal-oriented nature of human mobility networks.
However, the point of our contribution is not to dismiss the benefits of using betweenness centrality. Indeed, it is widely used to identify the most important nodes in a transportation network. Specifically, it identifies nodes that minimize overall travel distance across the network~\cite{ducruet_mesures_nodate}. However, it also corresponds to the potential bottlenecks and critical intersections for traffic flow management. Overall, it depicts the main structural vulnerabilities of a network~\cite{li_betweenness_2010}. It can even be used as a static predictor of congestion and traffic load~\cite{kirkley_betweenness_2018}.

Notwithstanding, researchers argue that information in networks does not exclusively follow shortest path, especially geodesic ones~\cite{newman_measure_2005,freeman_centrality_1991,stephenson_rethinking_1989}. Moreover, betweenness centrality is a static measure and may not capture dynamic changes in traffic conditions or temporal variations in network usage.

In contrast, the proposed Stigmergic Index offers an alternative perspective to traditional static and deterministic centrality measures. By shifting from rigid geodesic assumptions to decentralized, stochastic agent-based exploration with bounded rationality, this framework introduces behavioural variability into network analysis. Initial observations suggest it can help highlight certain structural patterns and flow distributions that complement conventional topological models.

One legitimate concern regarding the unconstrained stigmergic accumulation is the absence of a negative feedback mechanism. In classical traffic assignment, edge saturation increases travel time and triggers route diversion. However, the purpose of the Stigmergic Index in this baseline formulation is not to simulate dynamic real-time traffic assignment, but rather to diagnose inherent structural attraction and vulnerability to congestion.

Just as classical geodesic betweenness ignores capacity constraints while serving as a standard proxy for network centrality, our index measures the spontaneous funnelling of flows driven by spatial heading and collective memory before capacity-induced redistribution occurs. Interestingly, this self-reinforcing polarization accurately mirrors modern traffic phenomena---such as GPS-induced herding and arterial preference~\cite{acemoglu_informational_2018}---where drivers make navigation choices with regard to their perception of the network structure~\cite{parthasarathi_network_2013}. This can lead to problems such as Braess's Paradox~\cite{braess_paradox_2005}.

Future work will incorporate density-dependent negative feedback, inspired by ant collision mechanics under crowded conditions. Ethological observations of bottlenecks~\cite{dussutour_collective_2006} show that flow disruptions occur when stigmergic feedback becomes uncontrolled. By implementing greater \textit{Biorationality}---proximity to biological functions~\cite{madmar_unravelling_2026}---through modified ACO incorporating private-collective information synergy~\cite{czaczkes_synergy_2011} and congestion-avoidance strategies~\cite{poissonnier_experimental_2019,guerrieri2025}, we can assess the index's variability based on individual and collective behaviours.

Despite these initial results, several methodological steps remain necessary before the Stigmergic Index can be validated as a potential diagnostic tool. First, a comprehensive sensitivity analysis is required to evaluate how key parameters impact topological convergence and flow polarization. Second, incorporating edge capacity constraints is necessary to capture congestion dynamics and flow redistribution. We will evaluate the relevance of this indicator using real-life traffic data, and benchmark it against Randomized Shortest Paths~\cite{kivimaki_two_2016} and communicability measures~\cite{estrada_communicability_2008}.

%
%

\end{document}